\documentclass[journal]{IEEEtran}
\usepackage{framed,multirow}
\usepackage{amssymb}
\usepackage{latexsym}
\usepackage{url}
\usepackage{xcolor}
\usepackage{graphicx}
\usepackage{subfigure}
\usepackage{booktabs}
\usepackage{float}
\usepackage{amsmath}
\usepackage{bm}

\usepackage{cite}
\ifCLASSINFOpdf
\usepackage{amsmath}
\usepackage{array}
\usepackage[caption=false,font=normalsize,labelfont=sf,textfont=sf]{subfig}
\begin{document}
\title{Auto-Encoding Score Distribution Regression for Action Quality Assessment}

\author{Boyu~Zhang,
        Jiayuan~Chen,
        Yinfei~Xu,
        Hui~Zhang,
        Xu~Yang,
        Xin~Geng
       
 \thanks{Boyu Zhang and Jiayuan Chen contribute equally.}
 \thanks{\emph{Corresponding authors: Yinfei Xu; Hui Zhang.}}
 \thanks{Boyu Zhang and Jiayuan Chen are with the School of Computer Science and Engineering, Southeast University, Nanjing 211189, China (email: byz@seu.edu.cn; jiaychen@seu.edu.cn).}
 \thanks{Yinfei Xu is with the School of Information Science and Engineering, Southeast University, Nanjing 210096, China (email: yinfeixu@seu.edu.cn).}
 \thanks{Hui Zhang is with the Inspur Acadaemy of Science and Technology, Jinan 250000, China (email: zhanghui@inspur.com).}
 \thanks{Xu Yang and Xin Geng are with the School of Computer Science and Engineering, and the Key Lab of Computer Network and Information Integration (Ministry of Education), Southeast University, Nanjing 211189, China (email: s170018@e.ntu.edu.sg; xgeng@seu.edu.cn).}
 
}

\maketitle

\begin{abstract}
The action quality assessment (AQA) of videos is a challenging vision task since the relation between videos and action scores is difficult to model. Thus, AQA has been widely studied in the literature. Traditionally, AQA is treated as a regression problem to learn the underlying mappings between videos and action scores. But previous methods ignored 
data uncertainty in AQA dataset. To address aleatoric uncertainty, we further develop a plug-and-play module Distribution Auto-Encoder (DAE). Specifically, it encodes videos into distributions and uses the reparameterization trick in variational auto-encoders (VAE) to sample scores, which establishes a more accurate mapping between videos and scores. Meanwhile, a likelihood loss is used to learn the uncertainty parameters. We plug our DAE approach into MUSDL and CoRe. Experimental results on public datasets demonstrate that our method achieves state-of-the-art on AQA-7, MTL-AQA, and JIGSAWS datasets. Our code is available at \textit{https://github.com/InfoX-SEU/DAE-AQA}.

\end{abstract}

\begin{IEEEkeywords}
Action Quality Asessment, Uncertainty Learning,  Video Captioning
% Action Quality Asessment \sep Distribution Regression \sep Auto-Encoder
\end{IEEEkeywords}

% \fi
\IEEEpeerreviewmaketitle

\section{Introduction}

Action quality assessment (AQA) refers to the automatic scoring of behaviors in the video, such as scoring diving/gymnastics movements, and comparing which doctor has a higher surgical level. It is gaining increasing attention for its wide applications like the judgment of accuracy of an operation~\cite{Authors13} or score estimation of an athlete’s performance (at the Olympic Games)~\cite{Authors11,Authors14,Authors2019}. AQA analyzes how well an action is carried out, so it is more challenging than the traditional video action recognition (VAR) problem since identifying the difference in the same action category is imperceptible.  

Finding a solid link between the action score and videos is essential for AQA. In the past years, many AQA approaches ~\cite{Authors15,Authors16,Authors2019,core} tried to treat AQA as a regression problem and learned the direct mapping between videos and action scores. These works use the 3D convolutional neural network or LSTM to extract video features, and then regression methods are applied to get the prediction score. 
\begin{figure}[t]
\begin{center}
\includegraphics[width=0.95\linewidth]{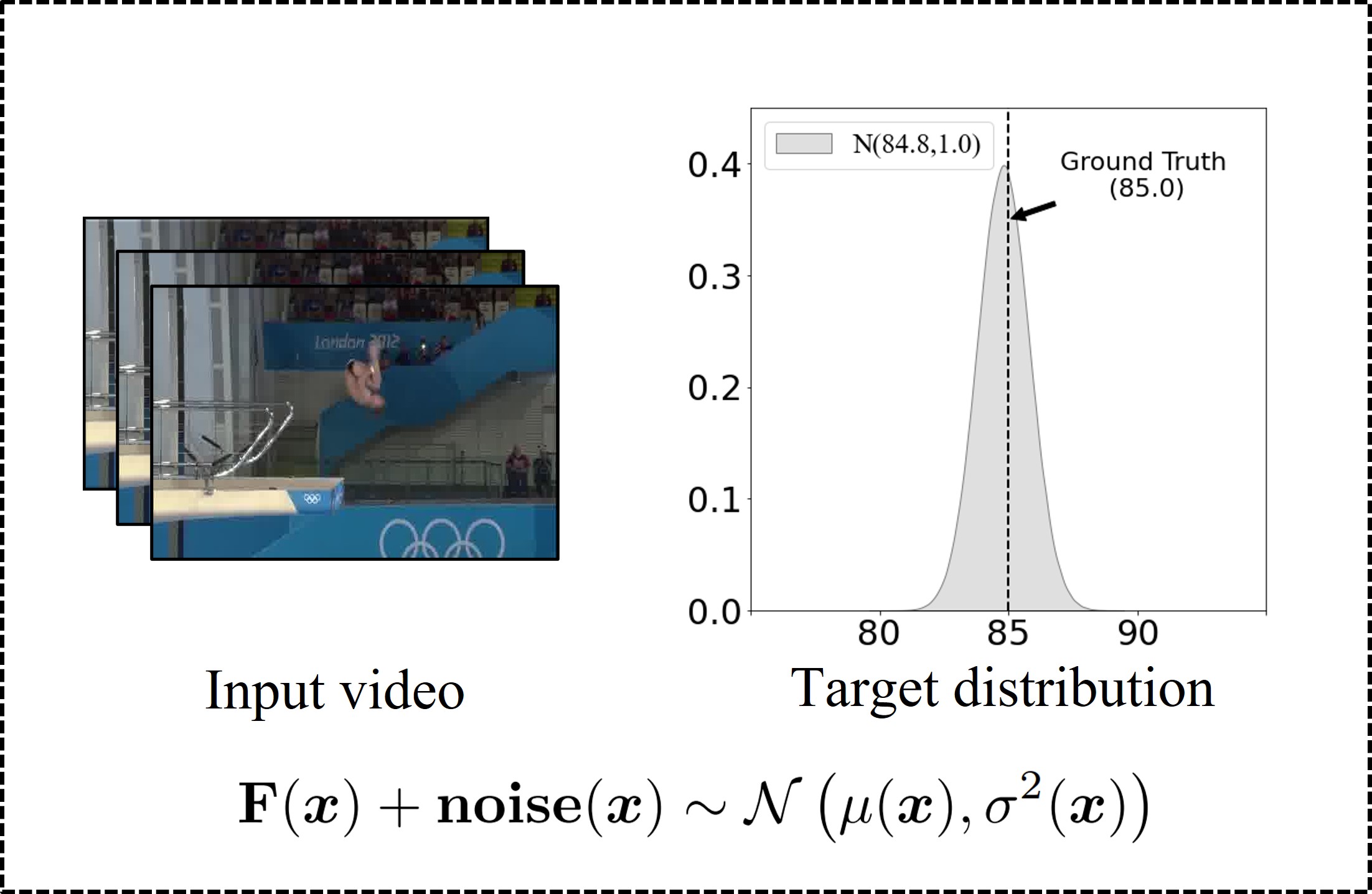}
\end{center}
\caption{Different actions have dissimilar aleatoric uncertainty when scored by the judges. Aleatoric uncertainty in DAE can be modeled as specific target distribution. e.g. Gaussian distribution. Our DAE achieves better performance since it can learn the adaptive variance from the data.}

\label{fig:1}
\end{figure}

In fact, most existing AQA methods ignore the inherent aleatoric uncertainty in datasets. AQA datasets are constructed subjectively by judges, which means there is observational noise that corrupts the target values. Thus, there does not exist an accurate mapping ${y}\sim \mathbf{F}(\boldsymbol{x})$ between label ${y}$ and data $\boldsymbol{x}$. Revised mapping form can be interpreted as
\begin{equation}
{y} = \mathbf{F}(\boldsymbol{x})+\mathbf{noise}(\boldsymbol{x})
\end{equation}

Given the above consideration, it is necessary to
model the observational $\mathbf{noise}(\boldsymbol{x})$ from a statistical point of view. To this end, we introduce uncertainty learning into AQA and propose a new regression model, named Distribution Auto-Encoder (DAE). By using DAE, the video features are synthesized into a score distribution. The final predicted score is then sampled from this distribution by reparameterization trick~\cite{Authors1}. Compared with the traditional regression method, our DAE method can automatically generate the inherent target distribution of 
videos. In this way, our approach can achieve a better prediction performance. Taking the Gaussian distribution as an example, which is shown in Figure~\ref{fig:1}, the action score predicted by DAE is continuously varying, and the variance of score distribution is adaptively learned from the dataset. %The AQA datasets include single-task datasets and multi-task %datasets. In recent years, fusion-based learning methods have %become more and more popular on multi-task datasets. To address our 5approach's effectiveness on both datasets, We also consider an %extension of DAE approach from single-task datasets to multi-task %datasets. In this paper, We envisage two models:

To illustrate the efficacy of our proposed method, We first construct our DAE model based on multi-layer perceptron (MLP). DAE-MLP requires score and video one-to-one information. The structure of DAE-MLP is composed of a features extractor and an encoder. Firstly, action videos are fed into an Inflated 3D ConvNets I3D~\cite{Authorsi3d} to extract feature vectors. Then, the feature vector is coded as a Gaussian distribution through the encoder, and the final predicted scores are sampled from this distribution.

Then we plug DAE into MUSDL~\cite{Authors2020} and CoRe~\cite{core} to show our approach is pluggable and effective. DAE-MT applies to multi-task datasets. It is proposed to utilize multiple tasks better. The feature extractor of DAE-MT is the same as DAE-MLP. In encoder part, DAE-MT predicts seven scores from seven judges rather than the final scores. The final score is the multiplication of difficulty degree(DD) and the raw score. Raw score is the average of seven judges' scores. DAE-CoRe adds DAE module to the last layer of the regression tree in CoRe. This method uses the regression tree to make predictions at smaller intervals.  

The main contributions of this work can be summarised as follows:
\begin{itemize}
    \item  We proposed a new plug-and-play regression module DAE to map video features into a score distribution inspired by VAE. DAE can address problems that previous works ignored aleatoric uncertainty during training. It provides an efficient solution to learn uncertainty and fit data more accurately.

    \item  A novel loss function is proposed to control the training of DAE. Specially, the proposed loss function can be weight-added. Using two weight-added parts together improves the performance of DAE.
    
    \item  Extensive model analysis experiments are performed on public databases to demonstrate that the proposed method achieves state-of-the-art performance in terms of the Spearman's Rank Correlation. And pluggable DAE only needs a slight time cost increment to improve the performance of baseline methods.

\end{itemize}

\section{Related Work}
Our work is closely related to three topics: action quality assessment,  uncertainty learning, and auto-encoder. In this section, we briefly review existing methods related to these topics.

\subsection{Action Quality Assessment}
Action Quality Assessment(AQA) automatically scores the quality of actions by analyzing features extracted from videos and images. It's different from conventional action recognition problems [11]–[16]. In the past few years, much work has been devoted to different AQA tasks, such as healthcare~\cite{Authors20}, sports video analysis~\cite{Authors11,Authors14,Authors2020}, and many others~\cite{Authors18,Authors30,Authors47,8636161}. The earliest model based on deep learning was proposed by Parmar {\em et~al.}~\cite{Authors14}, who use C3D-SVR and C3D-LSTM to predict Olympic scores. Based on the assumption that the final score is a set of continuous sub-action scores, the incremental label training method is introduced to train the LSTM model. Xiang {\em et~al.}~\cite{Authors17} 
choose to decompose video clips into action-specific clips and fuse the average features of clips to replace over full videos. More recently, Parmar {\em et~al.}~\cite{Authors2019} propose a C3D-AVG-MTL approach to learn Spatio-temporal features that explain three related tasks-fine-grained action recognition, commentary generation, and estimating the AQA score. Meanwhile, they collect a new multi-task AQA (MTL-AQA) dataset on a larger scale. Tang {\em et~al.} notice the underlying ambiguity of action scores. To address this problem, they propose an improved approach: uncertainty-aware score distribution learning (USDL)~\cite{Authors2020} based on label distribution learning (LDL)~\cite{Authors28}.
Multi uncertainty-aware score distribution learning (MUSDL) ~\cite{Authors2020} is designed to fit the multi-task dataset. It uses judges' information in the dataset and treated every judge as a scoring model. Contrastive Regression (CoRe)~\cite{core}
uses a pairwise strategy to regress the relative scores with reference to another video. Although the above models more or less take into account the interference of data uncertainty, they do not measure data uncertainty and thereby reduce the impact of noise.

\begin{figure*}[h]
\begin{center}
%\fbox{\rule{0pt}{2in} \rule{.9\linewidth}{0pt}}
 \includegraphics[height=5.7cm]{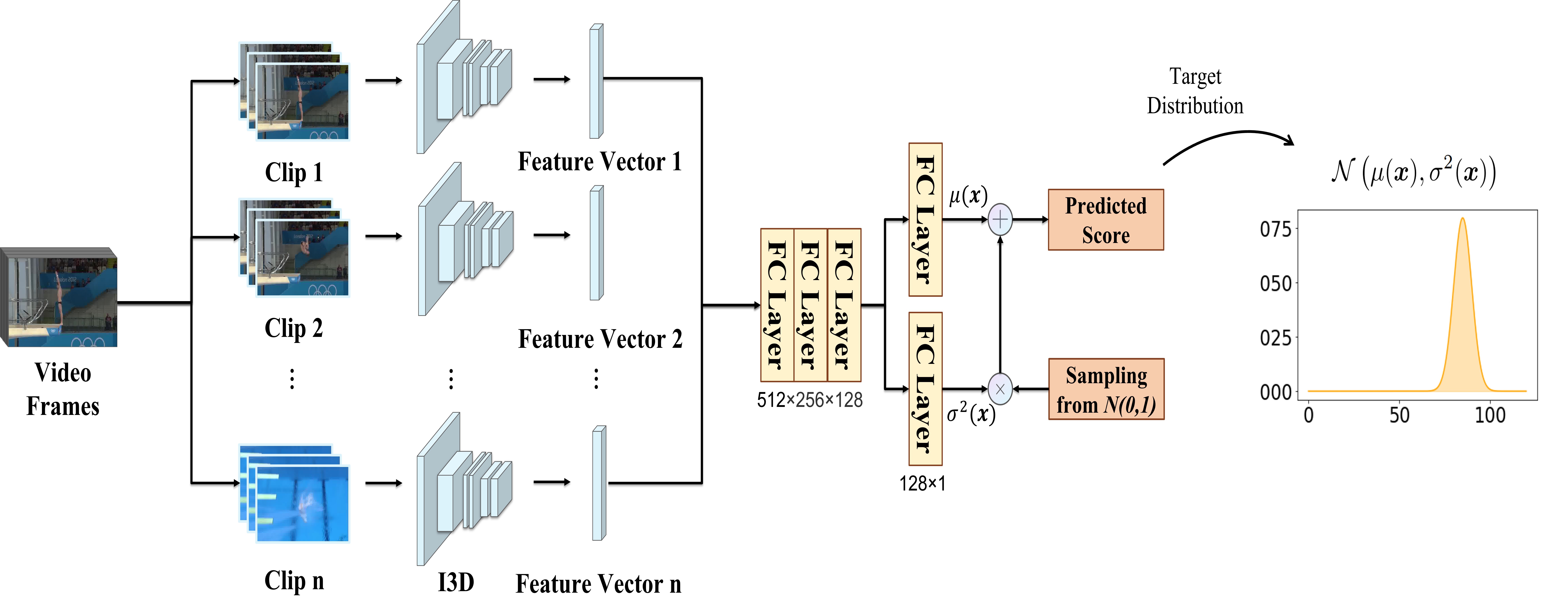} 
\end{center}
   \caption{The pipeline of DAE architecture contains two segments: video features extraction network and label distribution encoding network. 
   The input video is divided into $n$ small clips by down-sampling. Then the clips are sent into I3D ConvNets for extracting features. The final features are synthesized by three fully-connected layers.
   By Using an encoder, the video features are encoded into a Gaussian distribution, and the reparameterization trick is applied to generate samples from the distribution of the final predicted score. }
\label{fig:2}
\end{figure*}
\subsection{Uncertainty Learning}
Uncertainty study focuses on how to measure the implicit noise in a model or dataset. Two uncertainties are of major concern, epistemic uncertainty, and heteroscedastic aleatoric uncertainty. 
Epistemic uncertainty comes from the noise in model parameters or model outputs. Aleatoric uncertainty exists in dataset itself. Many people try to introduce uncertainty in modeling to get better results.~\cite{ul8,ul9}. Others also consider forming a generic learning paradigm to study uncertainty.
%plus
Geng {\em et~al.}~\cite{Authors28,Authors23} try to represent an instance by a specific distribution rather than one label or multiple labels. Pate {\em et~al.} ~\cite{ul31} use a risk level framework to measure uncertainty. Recently, uncertainty study on neural networks has been extensively studied~\cite{ul3,ul11,ul22}. And there are many tasks applied uncertainty analysis, {\em e.g.}. semantic segmentation ~\cite{ul19}, face recognition~\cite{ul2020} and object detection ~\cite{ul6,ul25,ul50} for the improvement of model robustness and effectiveness.

\subsection{Auto-Encoder}
The Auto-Encoder (AE) is first proposed by Hinton {\em et~al.}~\cite{AutoE}, which uses a multi-layer neural network to obtain low-dimensional expression of high-dimensional data. It uses the classical bottleneck network architecture and reconstructs the low-dimensional information back to the high-dimensional representation in the decoder. With the development of deep learning, many variant models based on AE have been created, such as Denoising Auto-Encoder for image denoising~\cite{AutoE2}, and Convolutional Auto-Encoder for image compression and feature extraction~\cite{AutoE3}.
According to the variational Bayes inference, the Variational Auto-Encoder (VAE)~\cite{Authors1} is proposed by Diederik P.Kingm and Max Welling based on the conventional auto-encoders. VAE used a unique reparameterization trick for sampling the hidden variable distribution~\cite{Authors1}. The method we proposed also refers to the Gaussian distribution encoding and sampling techniques in VAE. Specifically, our model maps video features to low-dimensional distributions and does not leverage the neural network for decoding, but directly outputs the final label through reparameterization.

\section{Distribution auto-encoder based on multi-layer perceptron (DAE-MLP)}

The distribution auto-encoder (DAE) is a plug-and-play regression module. In this section, We first constructed it based on multi-layer perceptron (MLP). DAE-MLP maps video clips to action score distribution via a deep neural network. The architecture of DAE-MLP is shown in Figure \ref{fig:2}. In particular, it consists of two parts, a video feature extractor to obtain video features (Section \ref{sec1}) and an auto-encoder for distribution learning (Section \ref{sec2}).

\subsection{Video Feature Extraction}
 \label{sec1}

Given an action video with $n$ frames, features need to be extracted first. As shown in the left half part of Figure \ref{fig:2}, the complete action video is down-sampled and divided into $n$ video clips,  \{$\boldsymbol{c}_1,...,\boldsymbol{c}_n$\}. Each video clip contains the same number of frames, representing a consecutive action snapshot. Parmar {\em et~al.}~\cite{Authors2019,Authors2020} pre-processed the video sequence in this way as well. The spatial resolution is improved by down-sampling since the number of network parameters can be reduced significantly. This technique was first proposed by Nibali {\em et~al.}\cite{Authors29} and widely used in the later works\cite{Authors2019,Authors2020}.

In the feature extraction of collected video clips, we use Inflated 3D ConvNets (I3D)~\cite{Authorsi3d}. I3D is a network model expanded from 2D to 3D convolution kernel, which has better performance in video analysis. I3D is followed by three fully connected layers, resulting in an $m$-dimensional feature.  Different clips share the exact weights of fully connected layers. After getting all the feature vectors of $n$ video clips,  we take the average as the final feature vector of the action video to guarantee that each video clip's information is considered equally.
 
\subsection{DAE-MLP}
 \label{sec2}

Compared with traditional regression methods, our approach captures aleatoric uncertainty. The action features are encoded into score distribution, and the final result is sampled from the auto-encoder output. This architecture enables learning a continuous distribution without loss in training procedure and quantifies the uncertainty of action score with high accuracy. The encoder uses a neural network to encode mean and variance simultaneously. The input 1024-dimensional feature vector $\boldsymbol{x}$ is encoded into  the parameters ${\mu(\boldsymbol{x})}$ and ${\sigma^{2}(\boldsymbol{x})}$ via a neural network.

Treating the action score as a random variable, we need to learn its score distribution and then sample the predicted score from the obtained distribution. For the input features, the first half of the structure of VAE is applied to encode the 1024-dimensional video feature $\boldsymbol{x}$ into a random variable ${y}$ through a probabilistic encoder ${p}({y}; \boldsymbol{\theta}(\boldsymbol{x}))$. The encoded random variable is assumed to be subject to Gaussian. distribution\footnote{Note that the Gaussian distribution is just one choice, and not a limitation in our method.}
\begin{equation}
{p}({y} ;\boldsymbol{\theta}(\boldsymbol{x}))=\frac{1}{\sqrt{2 \pi \sigma^2(\boldsymbol{x})} } \exp \left(-\frac{(y-\mu(\boldsymbol{x}))^{2}}{2 \sigma^{2}(\boldsymbol{x})}\right).
\label{eq11}
\end{equation}
 The parameters $\mu(\boldsymbol{x})$ and variance $\sigma^2(\boldsymbol{x})$ are used to quantify the quality and uncertainty of the action score. 
 
% and the reparameterization trick is then applied to sample from $y$ to output the predicted score.
 
 \noindent\textbf{Reparameterization Trick}: To generate a sample from Gaussian distributed $y$ as the predicted score and make full use of the two parameters in the score distribution at the same time, we invoke the reparameterization trick. 

According to reparameterization trick in VAE~\cite{Authors1}, assume that ${z}$ is a random variable, and ${z} \sim q({z}; \phi) $, $\phi$ is its parameter. We can express $z$ as a deterministic variable,  ${z}=g({\epsilon}; \phi) $, $\epsilon$ is an auxiliary variable with independent marginal $p({\epsilon})$,  and $g(\cdot;\phi)$ is a deterministic function parameterized by $\phi$. 

As illustrated in the right half part of Figure \ref{fig:2}, we do not directly sample from the score distribution, but firstly sample from $\epsilon$, which is distributed in $\mathcal{N}(0,1)$. Then, ${y}$ is calculated according to the sampling random variable $\epsilon$, mean parameter $\mu(\boldsymbol{x})$ and variance parameter $\sigma^2(\boldsymbol{x})$ of the auto-encoder output
\begin{equation}\label{eq:rp}
y=\mu(\boldsymbol{x})+\epsilon*\sigma(\boldsymbol{x}).
\end{equation} 

By applying the reparameterization trick, the score distribution sampling process is differentiable to ensure that the encoder training is feasible~\cite{Authors1}.

\begin{figure}[t]
\begin{center}
\includegraphics[width=1\linewidth]{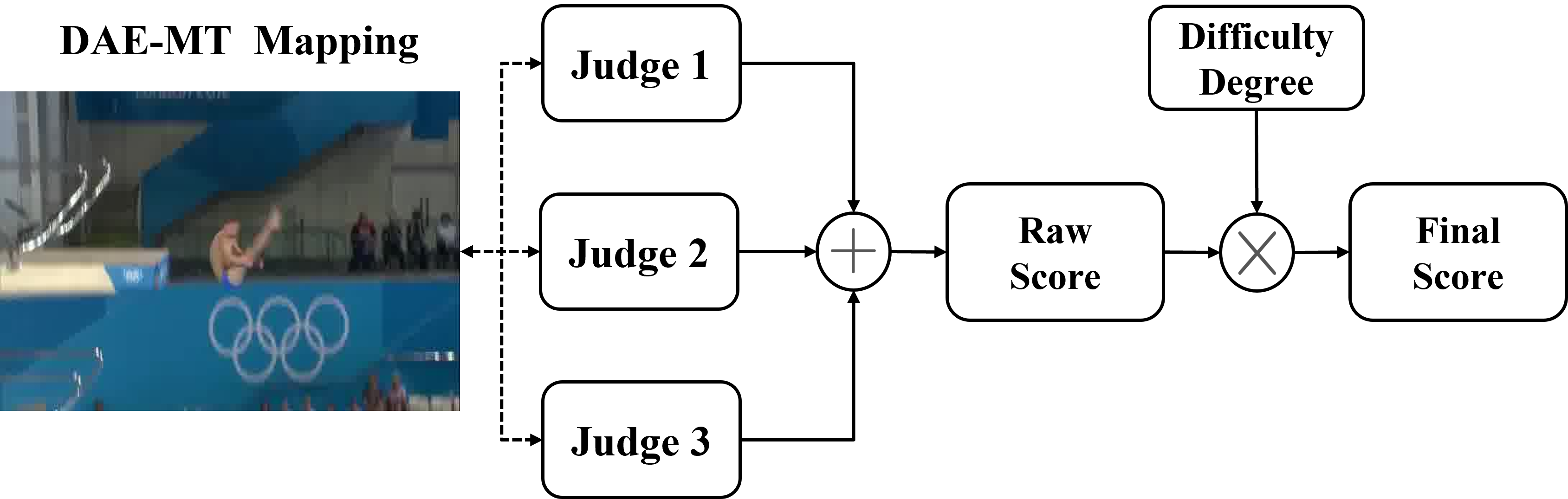}
\end{center}
\caption{The mapping rule of DAE-MT. DAE-MT uses the difficulty degree (DD) as prediction information, so it directly predicts the judges score distribution. DAE-MT outputs distributions for each judge. The final score is obtained by raw score multiplied by difficulty degree.}
\label{fig:3}
\end{figure}

\noindent\textbf{Loss Function}: Considering the optimization of parameters in neural networks from the perspective of likelihood, we obtain a form for loss function by expressing our goal as 
maximizing the log-likelihood of the target distribution \eqref{eq11},
\begin{equation}
\ln l(y ; \boldsymbol{x})=-\frac{1}{2} \ln (2 \pi)-\frac{1}{2} \ln \sigma^{2}(\boldsymbol{x})-\frac{(y-\mu(\boldsymbol{x}))^{2}}{2 \sigma^{2}(\boldsymbol{x})}
\end{equation}

The first term in RHS is a constant and can be ignored in maximization. Since maximizing a value is the same as minimizing the negative of that value, we interpret our overall loss function as
\begin{equation}
\begin{aligned}
\mathcal{L}&=\frac{1}{N} \sum_{i=1}^{N} \left[\frac{{\alpha}}{ \sigma\left(\mathbf{x}_{i}\right)^{2}}\left\|\mathbf{y}_{i}-\mu\left(\mathbf{x}_{i}\right)\right\|^{2}+{\beta} \log \sigma\left(\mathbf{x}_{i}\right)^{2}\right]\\
& \triangleq \frac{1}{N} \sum_{i=1}^{N} \alpha\mathcal{L}_{rec} + \beta\mathcal{L}_{sup}
\end{aligned}
\end{equation} 

$\alpha,\beta$ are the weights of two different parts of the reconstruction loss $\mathcal{L}_{rec}$ and the support loss $\mathcal{L}_{sup}$. The larger  $\alpha$ represents the more attention paid to uncertain information $\sigma^{2}(\boldsymbol{x})$. On the contrary, the larger $\beta$ represents DAE tends to become a traditional neural network regression model and pays more attention to regression precision.

\subsection{Uncertainty Regression via DAE }
We can compare DAE with traditional regression method to show the improvement by introducing uncertainty. Let $X$ = $(X_1,X_2,...,X_N), X_i$ = $(x_1,x_2,...,x_k)^T$ represent the input data, and ${Y}=(y_1,y_2,...,y_N)^T$ represents the label. The factors affecting $y$ are often multi-dimensional. Assume that there are $k$ factors $\mathbf{\bm{\beta}}=(\beta_{0},\beta_{1},\beta_{2} ,\cdots,\beta_{k})^T$ . The linear equation can be written as

\begin{equation}
y=\beta_{0}+\beta_{1} x_{1}+\beta_{2} x_{2}+\cdots+\beta_{k} x_{k}+\varepsilon
\end{equation}

After $N$ independent observations on $y$ and $x$, we can get $N$ groups of observations $(x_{i1},x_{i2},...,x_{ik}), \; i=1,2,...,N.$\footnote{$N \geq k +1$.}  It satisfy the following equations
\begin{equation}
y_{t}=\beta_{0}+\beta_{1} x_{i 1}+\beta_{2} x_{i 2}+\cdots+\beta_{k} x_{i k}+\varepsilon_{i}, \quad i=1, \ldots, n,
\end{equation}
where $\varepsilon_{1} \ldots \varepsilon_{N}$ are independent and with the same distribution as $\boldsymbol{\varepsilon}$. For regression problem ${Y}={X}^T \bm{\beta}+\boldsymbol{\varepsilon}$, we can solve it with traditional regression methods such as the least square method. Assuming that the error belongs to normal distribution $\boldsymbol{\varepsilon} \sim \mathcal{N}\left(0, \hat{\sigma}^{2}\right)$, the variance of estimation error~\cite{ul20} can be calculated by
\begin{equation}\label{eq:sse}
\hat{\sigma}^{2}=\frac{1-r^{2}}{N-2}\sum_{i=1}^{N}(y_i-\hat{y}_i)^{2},
\end{equation}
where $y_1,\hat{y}_1, \ldots, y_N,\hat{y}_N$ are the true and predicted labels, $r$ is correlation coefficient. 

Such estimation error is based on the assumption that the regression data $X$ is independent of the observational noise $\varepsilon$. However, due to the highly non-linearity of the prediction task at hand, the observational noise may be statistically coupled with the regression data. Traditional regression methods cannot estimate the noise variance while doing regression. Nevertheless, within neural uncertainty regressor (e.g., DAE), we can fit with data and estimate error specifically at the same time. The uncertainty should be related to $X$. We can write it as
\begin{equation}
\boldsymbol{\varepsilon} \sim \mathcal{N}\left(0, \sigma^{2}(X)\right)
\end{equation}

DAE allows us to learn underlying $\sigma^{2}(X)$ while regressing. And the training process of \eqref{eq:rp} can be regarded as an estimation of error. The prediction label of DAE is

\begin{equation}
Y \sim \mathcal{N}\left(X^T \bm{\beta}, \sigma^{2}(X)\right)
\end{equation}

Uncertainty regression introduced by DAE provides data augment and can be seen as a better fitting of underlying distribution in datasets.

\section{Plug-and-play Applications of DAE}

DAE is a generalization method and it can easily plug in any regression model. Plugging DAE in regression allows the baseline to capture aleatoric uncertainty. Under this consideration, we extend the plug-in application of DAE, we designed DAE-MT and DAE-CoRe on previous work MUSDL~\cite{Authors2020} and CoRe~\cite{core}, respectively.

\subsection{DAE-MT}

MUSDL~\cite{Authors2020} applies to multi-task datasets MTL-AQA. MUSDL shows that the scores from multiple judges and difficulty degree (DD) in MTL-AQA dataset are two essential pieces of information. Using them to calculate the final score of the player can be more reasonable. We adopt the same international diving scoring rules are adopted as MUSDL: Seven judges score, then remove the two lowest points and two highest points respectively. The remaining three scores are summed to get the raw score. The final score is obtained by multiplying the raw score and difficulty degree as follows,
\begin{equation}
S_{final}=(s_{max1}+s_{max2}+s_{max3})*DD.
\end{equation}

We plug DAE into MUSDL and proposed DAE-MT. Figure \ref{fig:3} shows the mapping of DAE-MT. We applied DAE-MT to predict the judges scores. It predicts scores without difficult degree (DD). Uncertainty only occurs when the judges score. So constructing the relationship between video and raw score directly is a more practical choice for learning uncertainty.

\begin{figure*}[t]
\begin{center}
\includegraphics[width=0.98\linewidth]{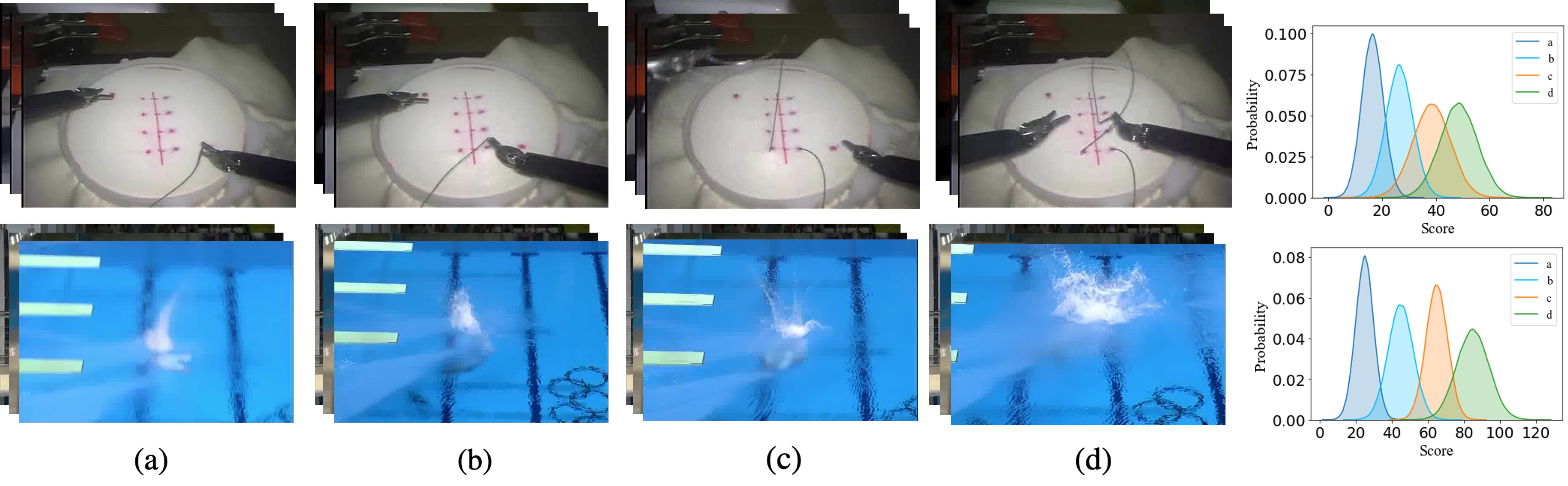}
\caption{Comparison of different distributions of different videos on MTL-AQA and JIGSAWS. Different operations indicate different prediction distributions. A small spray (a) will lead to a higher evaluation from judges, while a large spray (d) will get unsatisfactory scores.}
\label{fig:w1}
\end{center}
\end{figure*}

\subsection{DAE-CoRe}
CoRe~\cite{core} used a contrastive strategy to regress the relative scores between an input video and several exemplar
videos as references. It divides the range of the scores into several non-overlapping intervals and regresses in these small intervals with a binary tree. The final regression result can be written as
\begin{equation}
{y}=\mathcal{R}\left(\mathcal{I}_{\text {right }}-\mathcal{I}_{\text {left}}\right)+\mathcal{I}_{\text {left}}
\end{equation}
where $\mathcal{R}$ represents CoRe regressor, $\mathcal{I}_{\text {left}}$ and $\mathcal{I}_{\text {left}}$ represents the left and the right interval boundary respectively. 

DAE is easy to plug in the regressor of CoRe. We modify the last regression layer of the binary tree to capture aleatoric uncertainty. Regressor $\mathcal{R}$ can be modified from a neural network layer to DAE framework. This can be interpreted as
\begin{equation}
{y}=\mu\cdot\left(\mathcal{I}_{\text {right }}-\mathcal{I}_{\text {left}}\right)+\sigma \epsilon\cdot\left(\mathcal{I}_{\text {right }}-\mathcal{I}_{\text {left}}\right)+\mathcal{I}_{\text {left}}
\end{equation}

\section{Experiments}

\subsection{Datasets and Metrics}

\noindent\textbf{AQA-7~\cite{aqa7}:}
AQA-7 is a widely used sports dataset in AQA. It contains 7 kinds of sports videos: 370 samples from diving, 176 samples from gymnastic vault, 175 samples from skiing, 206 samples from snowboarding, 88 samples from synchronous diving - 3m springboard, 91 samples from synchronous diving - 10m platform and 83 samples from a trampoline. All the 1189 samples are divided into a training set of 863 samples and a testing set of 326 samples.

\noindent\textbf{MTL-AQA~\cite{Authors2019}:}
MTL-AQA contains 1412 diving samples, yet the data was comprised of 16 different events on diving to provide more variation. More specifically, MTL-AQA includes events of 10m Platform as well as 3m Springboard, both male and female athletes, individual or pairs of synchronized divers, and different views. The labels of MTL-AQA are also different from the previous diving dataset. The labels contain not only AQA scores from judges but also involve action class and commentary. In our experiment, we divide MTL-AQA into two parts, 1059 samples are used to train, and the remaining 353 samples are regarded as the testing set.

\noindent\textbf{JIGSAWS~\cite{Authorsj}:}
JIGSAWS is a surgical activity dataset for human motion modeling, consisting of 3 main tasks as “Suturing (S),” “Needle Passing (NP),” and “Knot Typing (KT).” Each task is divided into four folders. The video data of JIGSAWS are captured from a left camera and a right camera. Since these two videos are similar at a high level, we utilize the left to do our experiment. The label of JIGSAWS is surgical skill annotation, which is a global rating score (GRS) using modified objective structured assessments of technical skills approach.

\begin{figure}[t]
\begin{center}
\includegraphics[width=0.98\linewidth]{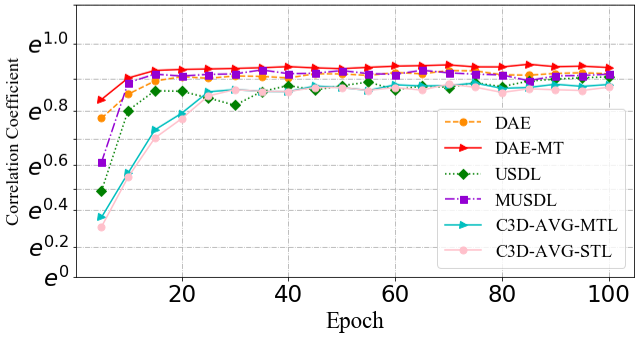}
\end{center}
\caption{Comparison of training procedure on MTL-AQA. The correlation coefficients are shown in exponential form.}
\label{fig:5}
\end{figure}

\noindent\textbf{Evaluation Metrics:} AQA uses Spearman’s rank correlation~\cite{Authors24} to measure the performance of our methods between the ground-truth and predicted score series.  Spearman’s correlation is defined as
\begin{equation}
\rho=\frac{\sum_{i}\left(p_{i}-\bar{p}\right)\left(q_{i}-\bar{q}\right)}{\sqrt{\sum_{i}\left(p_{i}-\bar{p}\right)^{2} \sum_{i}\left(q_{i}-\bar{q}\right)^{2}}}
\end{equation}

\begin{table*}[t]
\centering
\caption{Comparison of correlation coefficients on AQA-7. The best results are shown in bold in each block, respectively. }
\label{tab:aqa7}
\setlength{\tabcolsep}{4.5mm}{
\begin{tabular}{@{}c|cccccc|l@{}}
\toprule
Method   & Diving & Gym Vault & Skiing & Snowboard & Sync.3m & Sync.10m & Ave    \\ \midrule
Pose+DCT~\cite{dct} & 0.5300 & –         & –      & –         & –       & –        & –      \\
ST-GCN ~\cite{Authors24}  & 0.3286 & 0.5770    & 0.1681 & 0.1234    & 0.6600  & 0.6483   & 0.4433 \\
C3D-LSTM~\cite{Authors14} & 0.6047 & 0.5636    & 0.4593 & 0.5029    & 0.7912  & 0.6927   & 0.6165 \\
C3D-SVR~\cite{Authors14}  & 0.7902 & 0.6824    & 0.5209 & 0.4006    & 0.5937  & 0.9120   & 0.6937 \\
JRG ~\cite{jrg}     & 0.7630 & 0.7358    & 0.6006 & 0.5405    & 0.9013  & 0.9254   & 0.7849 \\
USDL ~\cite{Authors2020}   & 0.8099 & 0.7570    & 0.6538 & 0.7109    & 0.9166  & 0.8878   & 0.8102 \\
DAE-MLP  & \textbf{0.8420} & \textbf{0.7754}    & \textbf{0.6836} & \textbf{0.7230}    & \textbf{0.9237}  & \textbf{0.8902}   & \textbf{0.8258}      \\ \midrule
CoRe  ~\cite{core}   & 0.8824 & 0.7746    & \textbf{0.7115} & 0.6624    & 0.9442  & 0.9078   &0.8401 \\
DAE-CoRe & \textbf{0.8923} & \textbf{0.7786}    & 0.7102 & \textbf{0.6842}    & \textbf{0.9506}  & \textbf{0.9129}   &\textbf{0.8520}      \\ \bottomrule
\end{tabular}}
\end{table*}

\subsection{Implementation Details}

We implemented our DAE approach using the PyTorch framework. Two NVIDIA RTX 3090 GPUs are used to accelerate training. Moreover, We use 16 threads of Intel(R) Core(TM) i9-9900KF CPU @ 3.60GHz to accelerate data loading. 

Auto-encoder has three network layers. We choose \textit{ReLU} as activation function. The input layer's size is (1024,512), and the size of the hidden layers are (512,256) and (256,128). The final output layers of mean and variance are both (128,1). When evaluating DAE-MLP, the learning rate is $1e^{-4}$ on AQA-7, MTL-AQA and $5e^{-4}$, $2e^{-4}$ and $1e^{-4}$ on JIGSAWS-KT, -NP and -S, respectively. The hyper-parameters of DAE-MT and DAE-CoRe are set to be the same as the original baseline paper. According to our preliminary experiment, the weights of loss are selected as $\alpha=0.6$ and $\beta=0.4$. The optimizer is Adam~\cite{Authors21} on all datasets. On AQA-7 and MTL-AQA datasets, we selected the highest score as our method final performance during training. On JIGSAWS data, the final performance is averaged for the best ten consecutive scores to compare with previous methods, which is different from others.

\subsection{Experiment Results}
\noindent\textbf{AQA-7:}
We choose SOTA methods in the last five years to compare with our DAE, The final results are shown in Table \ref{tab:aqa7}. Our DAE-MLP method achieves 3.96\%, 2.43\%, 4.56\%,
1.70\%, 0.77\% and 0.83\% performance improvement for
each sports class compared with USDL~\cite{Authors2020} under Spearman’s correlation. The Average correlation rank of DAE-MLP improves 1.93\%. And the results of pluggable DAE also have achieved better performance in Diving (0.8923), Gym Vault (0.7786), Snowboard (0.6842), Sync.3m (0.9606), Sync.10m (0.9129) than CoRe~\cite{core}.

 \begin{table}[t]
 \caption{Comparison of correlation coefficients on MTL-AQA. The best results are shown in bold in each block, respectively. }
\label{tab:2}
 \begin{center}
\begin{tabular}{@{}l|l@{}}
\toprule
Method      & \qquad Sp. Corr.    \qquad    \qquad      \\ \midrule
C3D-SVR ~\cite{Authors14}    &\qquad 0.7716          \\
C3D-LSTM ~\cite{Authors14}   & \qquad0.8489          \\
MSCADC-STL ~\cite{Authors2019} \qquad \qquad\qquad &\qquad 0.8472          \\
MSCADC-MTL~\cite{Authors2019} \qquad \qquad \qquad & \qquad0.8612   \qquad       \\
C3D-AVG-STL ~\cite{Authors2019}\qquad& \qquad0.8960          \\
C3D-AVG-MTL~\cite{Authors2019} &  \qquad0.9044     \qquad     \\
USDL ~\cite{Authors2020}      & \qquad 0.9066          \\
DAE-MLP        & \qquad \textbf{0.9231} \\\midrule
MUSDL  ~\cite{Authors2020}     &  \qquad0.9273     \qquad     \\ 
DAE-MT        &  \qquad\textbf{0.9452} \qquad\\ \midrule
CoRe  ~\cite{core}     &  \qquad0.9512    \qquad     \\ 
DAE-CoRe        &  \qquad\textbf{0.9589} \qquad\\ 

\bottomrule
\end{tabular}
\end{center}

\end{table}
\noindent\textbf{MTL-AQA:}
We further applied DAE to MTL-AQA dataset to verify our approach's efficiency and show the comparison of differences between DAE and previous methods in more detail. Since multi judges information exists in MTL-AQA, we conduct an experiment using both DAE-MT and DAE-CoRe on MTL-AQA. The first block in Table \ref{tab:2} shows single-task training mode results. The prediction correlation coefficient of DAE-MLP reaches 0.9231, which is significantly improved in comparing the previous method. The second block shows that our method still has a good performance in multi-task mode, and the correlation coefficient of the final result is 0.9452, which exceeds the baseline model MUSDL~\cite{Authors2020}. And DAE-CoRe  (0.9589) also achieves a better result than CoRe (0.9512).

\noindent\textbf{JIGSAWS:}
The experiments results on JIGSAWS are shown in Table \ref{tab:1}. In all three surgical videos, our DAE-MT achieves a better performance of 0.73 (S), 0.71 (NP), 0.72 (KT), and 0.72 (Ave) than MUSDL. DAE-CoRe shows improvement in S (0.86) and KT (0.87), while getting the same score as CoRe in NP (0.86).

\begin{table}[t]
\begin{center}
\caption{Comparison of correlation coefficients on JIGSAWS. The best results are shown in bold in each block, respectively. }
\label{tab:1}
\begin{tabular}{@{}l|cccc@{}}

\toprule
Method \qquad& \multicolumn{1}{l}{S} & \multicolumn{1}{l}{NP} & \multicolumn{1}{l}{KT} &  \multicolumn{1}{l}{Avg. Corr.} \\ \midrule
 \small{ST-GCN ~\cite{Authors24}}& 0.31                  & 0.39                   & 0.58                   & 0.43                         \\
 \small{TSN ~\cite{Authors14}}   & 0.34                  & 0.23                   & 0.72                   & 0.46                         \\
 \small{JRG ~\cite{jrg}}  & 0.36                  & 0.54                   & \textbf{0.75}                   & 0.57                 \\
 \small{USDL~\cite{Authors2020}}  & 0.64                  & 0.63                   & 0.61                   & 0.63                        \\
  \small{DAE-MLP}    & \textbf{0.73}        & \textbf{ 0.72}            &   0.72              & \textbf{0.72}  
    \\ \midrule
    \small{MUSDL~\cite{Authors2020}} & 0.71                  & 0.69                   & 0.71                   & 0.70   \\
    \small{DAE-MT}    & \textbf{0.78}        & \textbf{0.74}            &   \textbf{0.74}              & \textbf{0.76} 
    \\\midrule
    \small{CoRe~\cite{core}} & 0.84                  & \textbf{0.86  }                 & 0.86                  & 0.85   \\
    \small{DAE-CoRe}    & \textbf{0.86}       & 0.86         &   \textbf{0.87 }             & \textbf{0.86} 
    \\
 \bottomrule
\end{tabular}
\end{center}
\end{table}

\begin{table}[t]
\begin{center}
\caption{Comparison of different sampling distributions.}
\label{tab:dis}
\begin{tabular}{@{}l|l@{}}
\toprule
Method    & \quad Sp. Corr.   \quad         \\ \midrule
Student't  Distributions&\quad 0.9399          \\
Laplace  Distributions& \quad0.9290          \\
Triangular Distributions &\quad 0.9207          \\
LogisticNormal Distributions\quad \qquad \qquad&\quad 0.9262          \\
Gaussian  Distributions& \quad\textbf{0.9452} \\
\bottomrule
\end{tabular}
\end{center}
\end{table}

\subsection{Analysis}

\begin{figure*}[t]
\centering
\subfigure[]{\includegraphics[height=5.2cm]{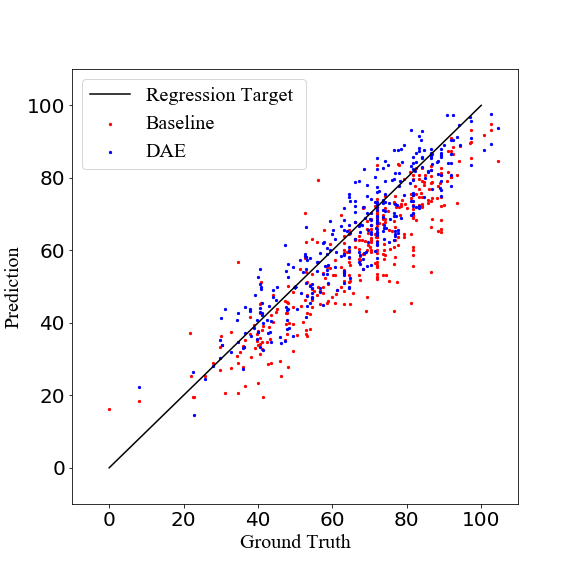}}
\subfigure[]{\includegraphics[height=5.2cm]{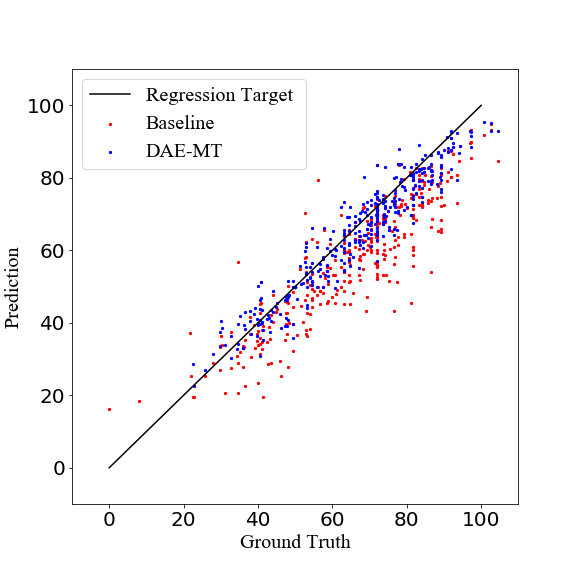}}
\subfigure[]{\includegraphics[height=5.2cm]{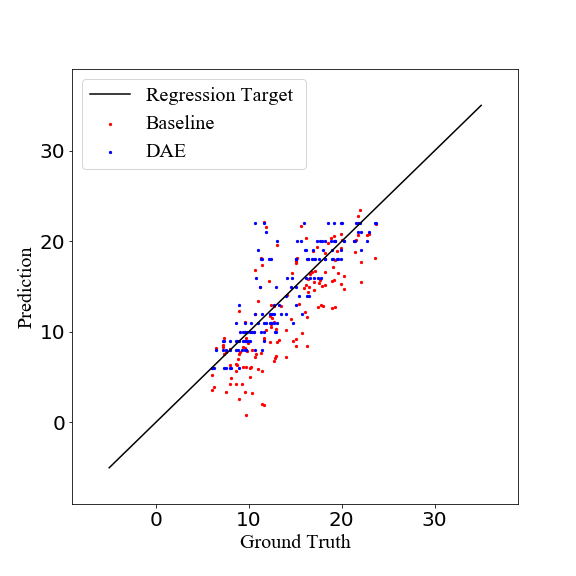}}
\subfigure[]{\includegraphics[height=5.2cm]{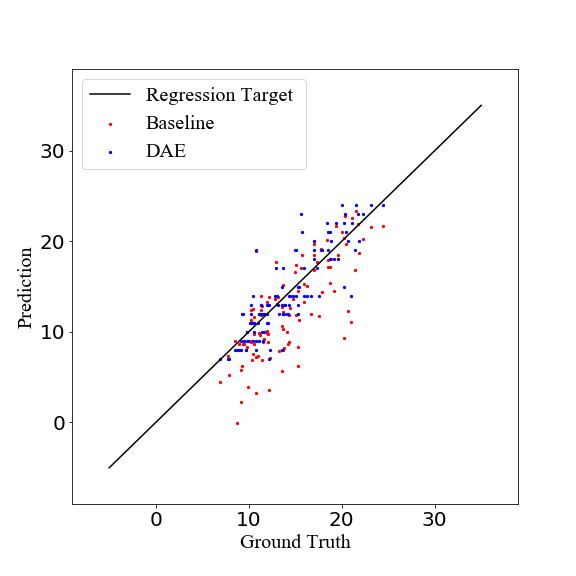}}
\subfigure[]{\includegraphics[height=5.2cm]{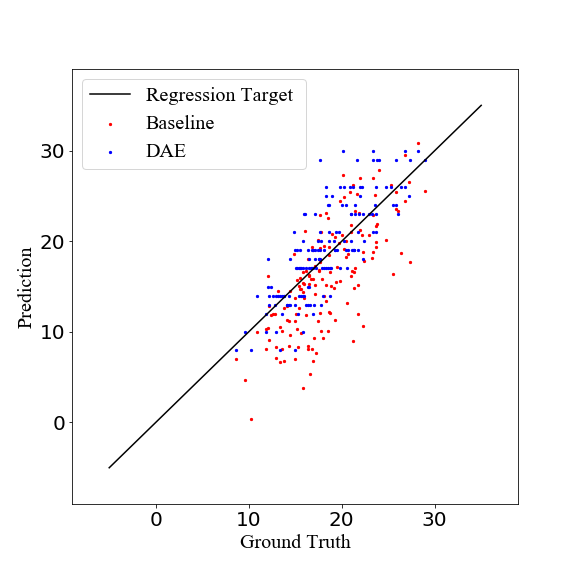}}
\caption{Comparisons of our methods and regression baseline.  (a) and (b) show the results of DAE and DAE-MT on MTL-AQA. (c), (d) and (e) show the results on JIGSAWS-KT, JIGSAWS-NP and JIGSAWS-S.}
\label{fig:reg}
\end{figure*}

\begin{figure}[t]
\begin{center}
\includegraphics[height = 6.5cm,width=0.75\linewidth]{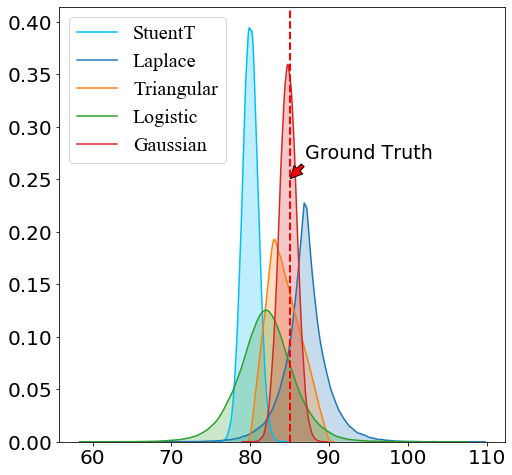}
\end{center}
\caption{Parallel experiments on different distributions. We tested five distributions of auxiliary variable $p(\boldsymbol{\epsilon})$ on MTL-AQA.}
\label{fig:distributio}
\end{figure}

\noindent\textbf{Study on Different Distributions:}
When our method encodes features into distributions, the form of distribution is not limited to normal distribution. For the reparameterization trick for any "location-scale" (Eq. \eqref{eq:rp}) family of distributions, we can choose the standard distribution (with location = 0, scale = 1) as the auxiliary variable ${\epsilon} \sim p({\epsilon})$, and let 
\begin{equation*}
    g(\cdot;\phi) = location + scale* \epsilon
\end{equation*}

We do parallel experiments on Laplace, Elliptical, Student’s t, Logistic, Uniform, Triangular and Gaussian distributions on MTL-AQA dataset. A case is shown in Fig\ref{fig:distributio} and the evaluation is summarized in Table \ref{tab:dis}. We find that Gaussian distribution performs best on this dataset. Although this result is largely in line with our expectations, it does not mean that Gaussian distribution is the most appropriate in other datasets or other video evaluation tasks. For specific datasets and applications, we need to select the best distribution through further experiments.
%\vspace{-1.0em}

\noindent\textbf{Case Study:}
We applied a case study to compare the prediction distribution of DAE for different videos on both MTL-AQA and JIGSAWS datasets, as shown in Figure \ref{fig:w1}. Each line shows four videos and their distributions predicted by our DAE. Different action quality generates different mean and variance distribution, which corresponds to the actual performance. Taking diving as an example, a small spray Figure \ref{fig:w1}(a) will lead to a higher evaluation of judges, while a large spray Figure \ref{fig:w1}(d) will get bad scores. This proves the effectiveness of our method. DAE can make an effective prediction according to different video content. The predicted distribution parameters are adaptive according to the video itself.
\footnote{Whether the variance is positive does not affect the reparameterization trick. Here variance shows in absolute value.}

\noindent\textbf{Regression Analysis:}
We compare our method and regression baseline detailedly with plotting to scatter diagrams. In Figure \ref{fig:reg}(a)-(b), we show the results of methods DAE-MLP and DAE-MT on MTL-AQA. In Figure \ref{fig:reg}(c)-(e), we show the results on JIGSAWS-KT, JIGSAWS-NP and JIGSAWS-S respectively. Regression target is the ideal regression result. The closer scatters are to this line, the better regression is. From Figure \ref{fig:reg}, we can see that DAE-MLP and DAE-MT both perform well. Scatters of our method are all closer to the target line than regression baseline. And the prediction of our method is more concentrated. 

\begin{figure}[t]
\begin{center}
\includegraphics[width=0.75\linewidth]{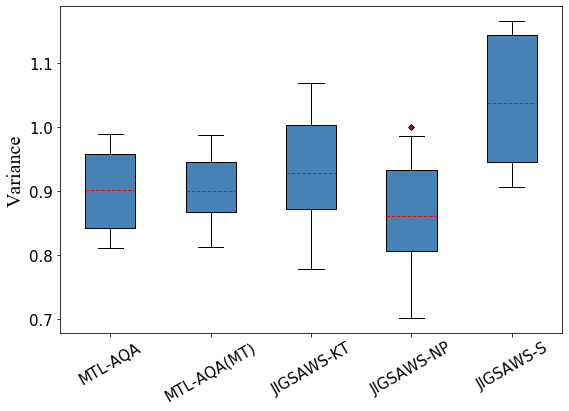}
\end{center}
\caption{Variation range of variance in parallel training on MTL-AQA and JIGSAWS.}
\label{box}
\end{figure}

We further compare the training of DAE with previous models USDL~\cite{Authors2020} , MUSDL~\cite{Authors2020} ,C3D-AVG-STL and C3D-AVG-MTL~\cite{Authors2019}. The comparison results on MTL-AQA dataset are shown in Figure \ref{fig:5}. It can be seen that the final stable correlation coefficient of DAE is higher than that of other methods, and DAE converges faster during training and there are minor fluctuations.

Besides, a parallel experiment is carried out to find out the variation range of variance.  We trained DAE-MLP twenty times to observe the variance of the best-performing model in each training round. Box chart Figure \ref{box} shows the results of the parallel comparison. The variation range of variance is $\pm0.1$ in MTL-AQA training rounds. And range in JIGSAWS training rounds is $\pm0.2$. From quartiles, it can be seen that variances in five observations are stable relatively. The upper and lower quartile lines are roughly the same distance from the median. There is only one outlier in rounds JIGSAWS-NP, it goes beyond the maximum observation and exceeds the upper edge. This may be caused by sampling.

\begin{table}[t]
\caption{Comparison of training results with different losses on MTL-AQA and JIGSAWS. We used the regression method as our baseline. We used DAE-MLP and DAE-MT training, respectively.}
\label{tab:loss}
\begin{center}
\begin{tabular}{@{}l|cccc@{}}
\toprule
Loss
  & \multicolumn{1}{l}{Sp. Corr.} & \multicolumn{1}{l}{Sp. Corr.(MT)} \quad \\ \midrule
Regression baseline\qquad  & 0.8905                  &   0.8905                                     \\

MSE loss    & 0.9180                 & 0.9415                                      \\
Sup+Rec loss   &\textbf{0.9231}                 & \textbf{0.9452}                              \\
\toprule
Loss
  & \multicolumn{1}{l}{Sp. Corr.} & \multicolumn{1}{l}{Sp. Corr.(MT)} \quad \\ \midrule
Regression baseline  & 0.68                  &   0.68                                     \\
MSE loss    & 0.71                & 0.72                                    \\
Sup+Rec loss   & \textbf{0.72}                 & \textbf{0.76}                              \\
\bottomrule
\end{tabular}
\end{center}
\end{table}

\noindent\textbf{Loss Study:}
A two parts likelihood loss function is adopted in this paper. We evaluated the effectiveness of our loss function in this section. We used common MSE loss $ \mathcal{L}_{MSE}=\frac{1}{N} \sum_{i=1}^{N}\left(y_{i}-(\mu(x_{i})+\sigma^2(x_{i}))\right)^{2}$for comparison.

Table \ref{tab:loss} shows ablation results on MTL-AQA and JIGSAWS. We first used the single-task training mode for the loss experiment on MTL-AQA. When common MSE loss is used, the mean of predicted distribution will be more likely to close the final predicted score. At this point, the learning of variance is reduced. Our regression result is similar to the baseline of USDL. Their baseline scores reached 0.8905, and our scores reached 0.9091. When we used our likelihood loss, we find that the training performance has greatly improved. At the same time, the stability of variance has also been improved after many observations. The performance of DAE-MT in multi-task training mode has similar trends to DAE. The performance reached 0.9415 when MSE loss is used. When training with our loss, DAE-MT achieved the best performance with reaching a score of 0.9452. Compared with baseline, the performance of our best method increases by 6.1$\%$. The second block in Table \ref{tab:loss} shows ablation results on JIGSAWS. It can be seen that The effect of combined loss is better than that of a single loss and baseline also.

\noindent\textbf{Inference Time:}
At inference time, all methods are applied to ten video samples at a time, Table \ref{table:time} shows the average inference time on MTL-AQA. We tested all the models using a GPU, NVIDIA RTX 3090. It shows that only a slight time increment is needed to plug DAE into other regression methods. DAE-MT and DAE-CoRe involve 7\%  and 9.7\% inferencing time increments, respectively.

\begin{table}[t]
\centering
\caption{Comparison results of inference time.}
\setlength{\tabcolsep}{2mm}{
\begin{tabular}{cccccc}
\hline
Method    & DAE-MLP & MUSDL & DAE-MT & CoRe & DAE-CoRe\\ \hline
Time(ms) & 0.84   & 1.42 & 1.52 & 2.16   & 2.37    \\ \hline
\end{tabular}}
\label{table:time}
\end{table}

\begin{figure*}[t]
\begin{center}
\includegraphics[width=1\linewidth]{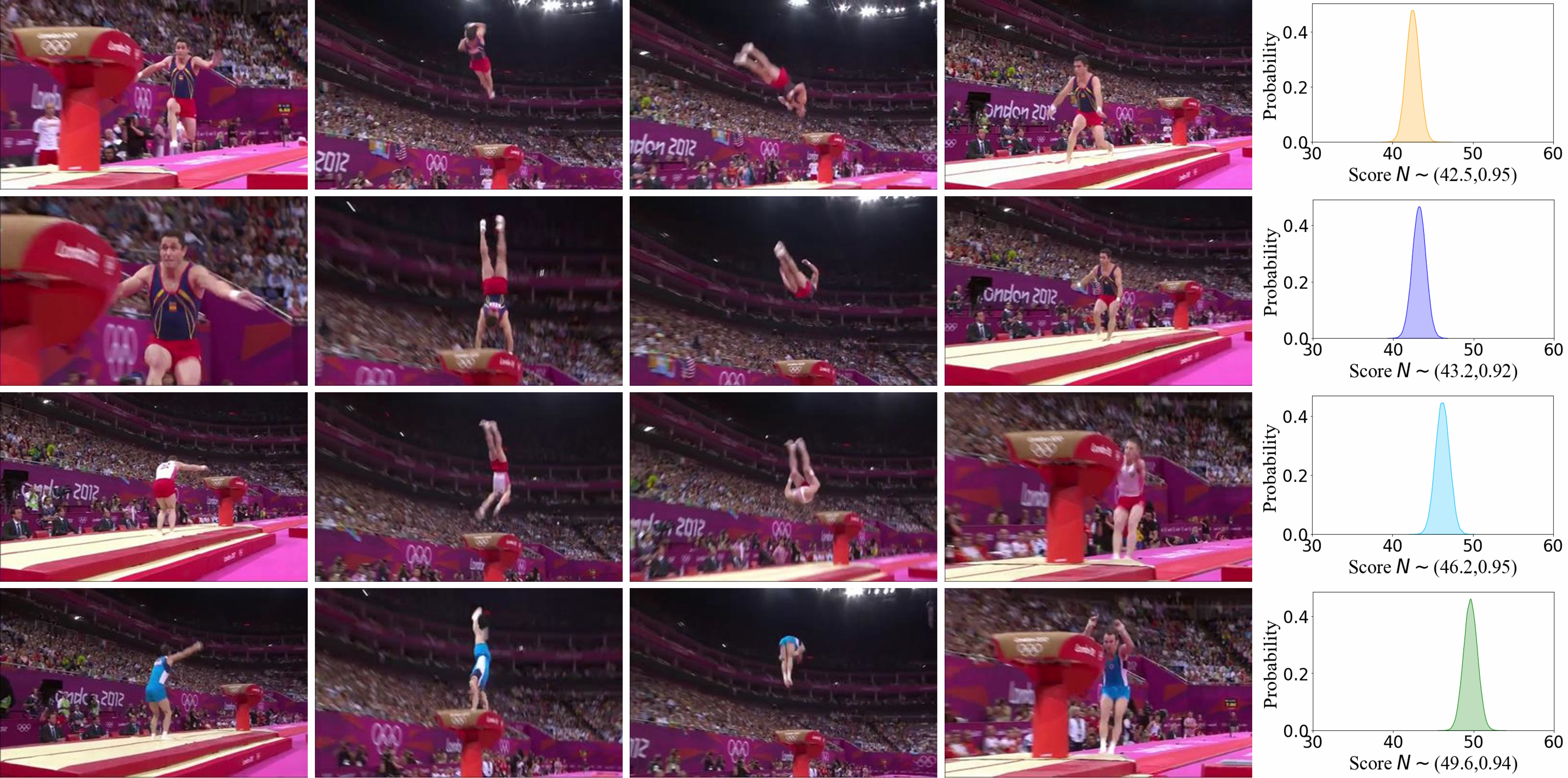}
\caption{Case study on AQA-7 (Gym Vault) dataset. Each row indicates four frames of a video corresponding to its prediction distribution.}
\label{fig:show1}
\end{center}
\end{figure*}

\begin{figure*}[h]
\begin{center}
\includegraphics[width=1\linewidth]{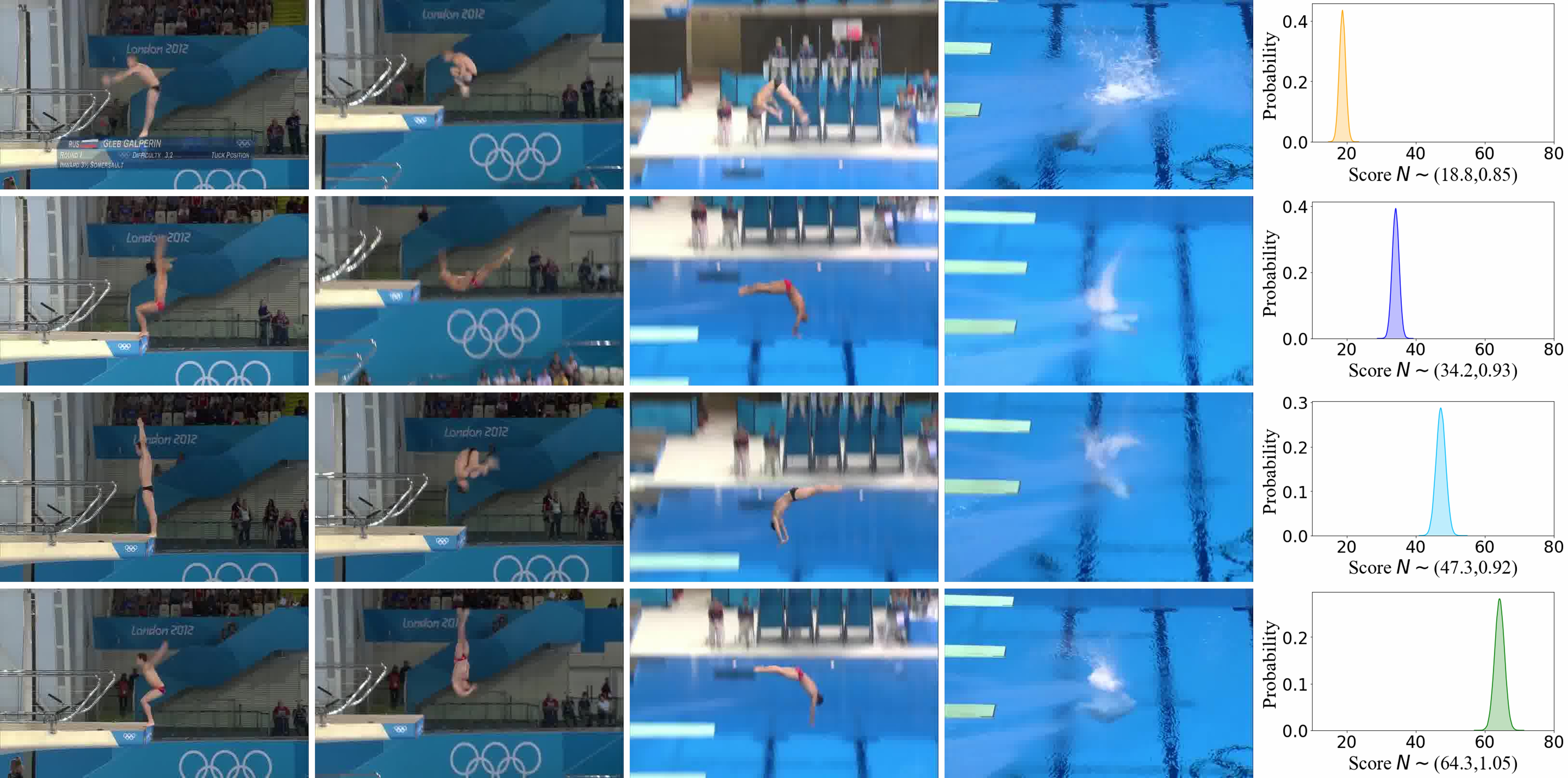}
\caption{Case study on MTL-AQA dataset. Each row indicates four frames of a video corresponding to its prediction distribution.}
\label{fig:show2}
\end{center}
\end{figure*}

\section{Conclusion}
In this paper, we propose a new method for action quality assessment.  Referring to the architecture of variational auto-encoders, we propose a new regression model, Distribution auto-encoders (DAE). Furthermore, DAE is pluggable and can be extended on any regression method. We have tested our approach on AQA-7,MTL-AQA, and JIGSAWS to show that our method outperforms better than the state-of-the-art. The increment time cost of inferencing is also acceptable. Although this paper is explicitly geared toward action quality assessment,  DAE provides a general solution paradigm for uncertainty learning. A specific distribution is used to represent an instance (or features), the parameters of the distribution are obtained by the encoder to quantify the value of the label, and the uncertainty is quantified by sampling from the distribution. This method captures the uncertainty on the basis of the traditional regression method and learns the inherent characteristics of uncertainty through the multi-layer neural networks, which is more explainable in theory. In the future, we plan to apply this method to other video analysis problems, such as age estimation~\cite{Authors26,Authors43} and facial beauty prediction~\cite{Authors33,Authors49}.

\appendices
 \section{Case Study}

\begin{figure*}[h]
\begin{center}
\includegraphics[width=1\linewidth]{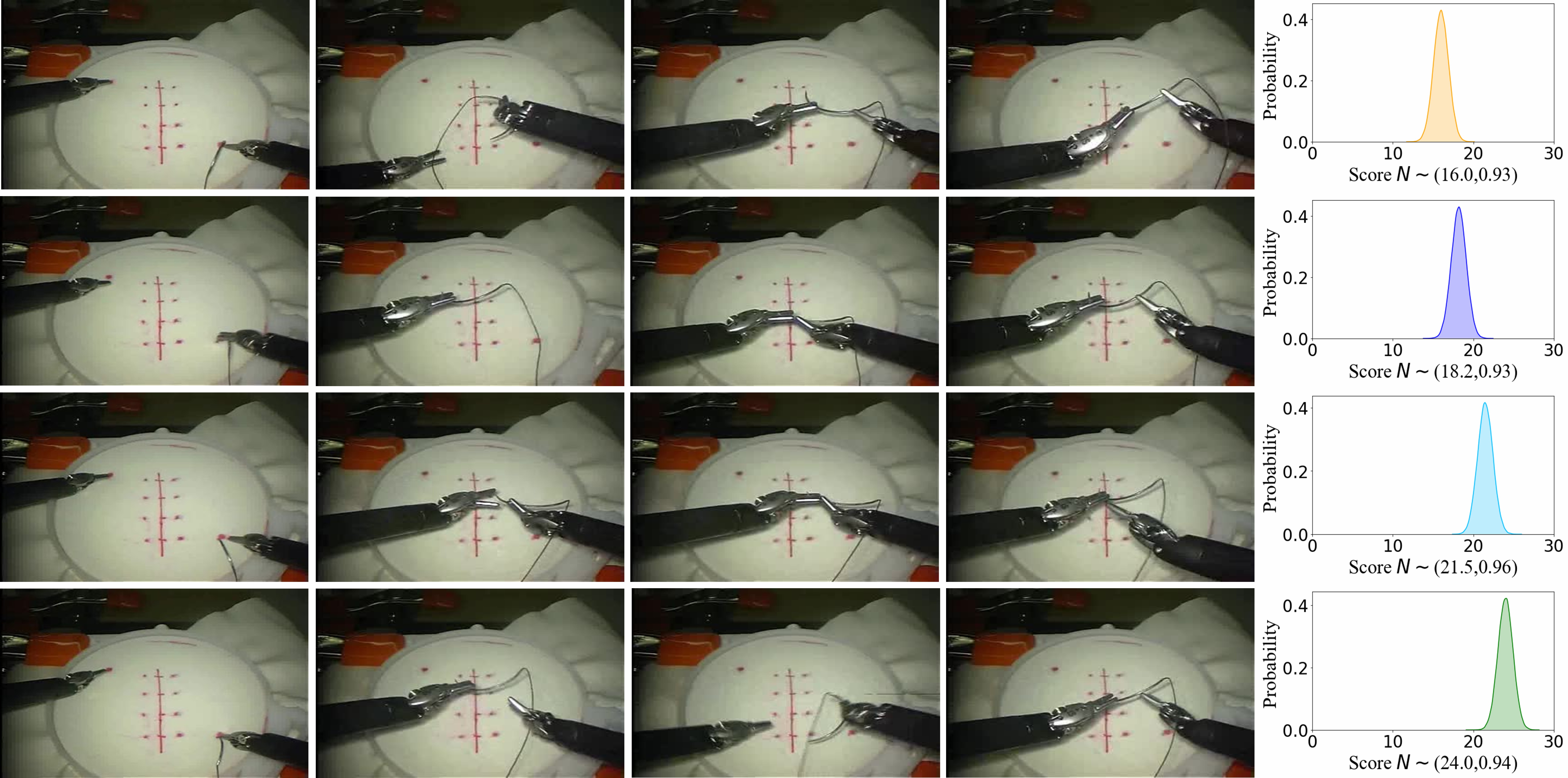}
\caption{Case study on JIGSAWS (KT) dataset. Each row indicates four frames of a video corresponding to its prediction distribution. }
\label{fig:show3}
\end{center}
\end{figure*}

\begin{figure*}[t]
\begin{center}
\includegraphics[width=1\linewidth]{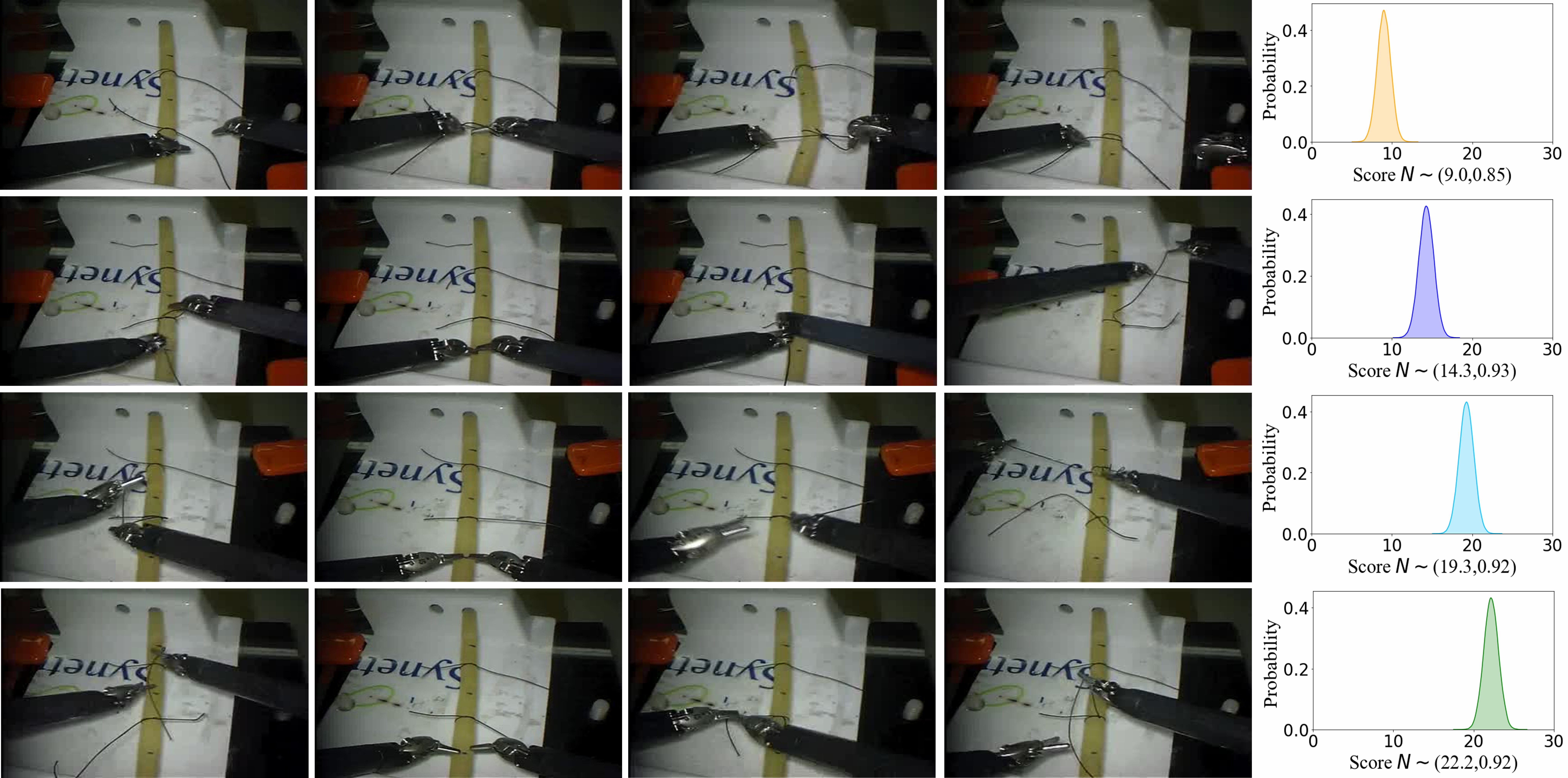}
\caption{Case study on JIGSAWS (NP) dataset. Each row indicates four frames of a video corresponding to its prediction distribution. }
\label{fig:show4}
\end{center}
\end{figure*}

We conduct more case studies here. Figure \ref{fig:show1}-\ref{fig:show4} illustrate that DAE captures the uncertainty in video quality assessment datasets well. Uncertainty in each case can be learned adaptively by the case itself. Specifically, for an excellent movement, the variance of the evaluation is minor, and the judges' scores will be close to stable, while for low-quality movements, the range of changes in the score will increase, and the reliability will decrease accordingly.

% Note that often IEEE papers with subfigures do not employ subfigure
% captions (using the optional argument to \subfloat[]), but instead will
% reference/describe all of them (a), (b), etc., within the main caption.
% Be aware that for subfig.sty to generate the (a), (b), etc., subfigure
% labels, the optional argument to \subfloat must be present. If a
% subcaption is not desired, just leave its contents blank,
% e.g., \subfloat[].

% An example of a floating table. Note that, for IEEE style tables, the
% \caption command should come BEFORE the table and, given that table
% captions serve much like titles, are usually capitalized except for words

% or


% \appendices

% % you can choose not to have a title for an appendix
% % if you want by leaving the argument blank
% \section{}

% % use section* for acknowledgment
% \section*{Acknowledgment}


\bibliographystyle{IEEEtran}
\bibliography{IEEEabrv,IEEEexample}

\ifCLASSOPTIONcaptionsoff
  \newpage
\fi

%\newpage



% \end{thebibliography}


% \end{IEEEbiography}


%\newpage



%\vfill

\end{document}